\documentclass[10pt,twocolumn,letterpaper]{article}

\usepackage{cvpr}
\usepackage{times}
\usepackage{epsfig}
\usepackage{graphicx}
\usepackage{amsmath}
\usepackage{amssymb}
\usepackage{hyperref}
\usepackage{url}
\usepackage{dirtytalk}
\usepackage{caption}
\usepackage{subcaption}
\usepackage{color}
\usepackage{algpseudocode}
\usepackage{xcolor}
\usepackage{multirow}
\usepackage{wrapfig}
\usepackage{lipsum}
\usepackage{enumitem}
\usepackage[font=small,skip=0pt]{caption}
\usepackage[export]{adjustbox}
\usepackage[ruled]{algorithm2e}
\usepackage{pythontex}

\DeclareMathOperator*{\argminA}{arg\,min} 

\newcommand{\model}{ADP }

\def\cvprPaperID{} 

\ifcvprfinal\fi
\begin{document}

\title{Adversarial Data Programming: Using GANs to Relax the Bottleneck of Curated Labeled Data}

\author{Arghya Pal, Vineeth N Balasubramanian\\
Department of Computer Science and Engineering\\
Indian Institute of Technology, Hyderabad, INDIA\\
{\tt\small \{cs15resch11001,vineethnb\}@iith.ac.in}
}

\maketitle

\begin{abstract}
Paucity of large curated hand-labeled training data for every domain-of-interest forms a major bottleneck in the deployment of machine learning models in computer vision and other fields. Recent work (Data Programming) has shown how distant supervision signals in the form of labeling functions can be used to obtain labels for given data in near-constant time. In this work, we present Adversarial Data Programming (ADP), which presents an adversarial methodology to generate data as well as a curated aggregated label has given a set of weak labeling functions. We validated our method on the MNIST, Fashion MNIST, CIFAR 10 and SVHN datasets, and it outperformed many state-of-the-art models. We conducted extensive experiments to study its usefulness, as well as showed how the proposed ADP framework can be used for transfer learning as well as multi-task learning, where data from two domains are generated simultaneously using the framework along with the label information. Our future work will involve understanding the theoretical implications of this new framework from a game-theoretic perspective, as well as explore the performance of the method on more complex datasets. 
\end{abstract}
\section{Introduction}
\label{sec_intro}
\vspace{-3pt}
Curated labeled data is a key building block of modern machine learning algorithms, and a driving force for deep neural network models. The large parameter space of deep models requires very large labeled datasets to build effective models that work in practice. However, this inherited dependency on large curated labeled data has become the major bottleneck of progress in the use of machine learning and deep learning in computer vision and other domains \cite{sun2017revisiting}. Creation of large scale hand-annotated datasets in every domain is a challenging task due to the requirement for extensive domain expertise, long hours of human labour and time - which collectively make the overall process expensive and time-consuming. Even when data annotation is carried out using crowdsourcing (e.g. Amazon Mechanical Turk), additional effort is required to measure the correctness (or goodness) of the obtained labels. We seek to address this problem in this work \footnote{This paper is accepted in CVPR 2018}.

\begin{figure*}
    \centering
    \includegraphics[width=\textwidth,height=0.3\textwidth]{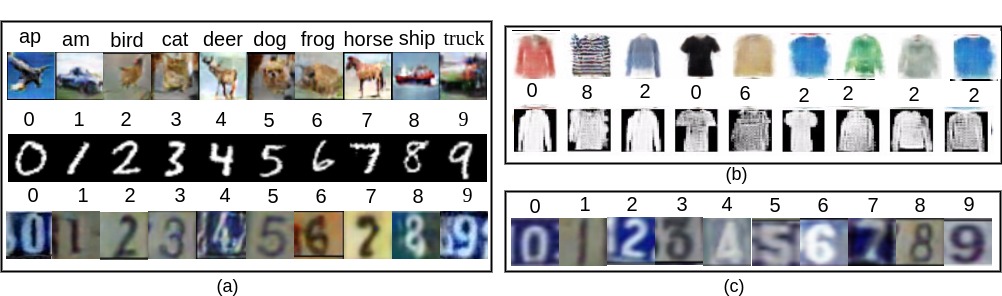}
    \caption{(a) Sample results of image-label pairs generated using the proposed \model framework trained on CIFAR-10, MNIST and SVHN datasets (top to bottom respectively). Note that the label is generated by our model; (b) Demonstration of cross-domain multi-task learning using \model, where the same model generates data from both Fashion MNIST and LookBook datasets (Section \ref{sec_analysis}). Note that Fashion MNIST is grayscale while LookBook is color, and the model still generates both data effectively; (c) Demonstration of transfer learning of our \model from MNIST dataset (source domain) to generate image-label pairs on the SVHN dataset (target domain).}
    \label{fig:Motivational_example}
\end{figure*}

In particular, we focus on automatically learning the parameters of a given joint image-label probability distribution (as provided in training image-label pairs) with a view to automatically create labeled datasets. To achieve this objective, we exploit the use of distant supervision signals to generate labeled data. These distant supervision signals are provided to our framework as a set of weak labeling functions which represent domain knowledge or heuristics obtained from experts or crowd annotators. Writing a set of labeling functions (as we found in our experiments) is fairly easy and quick, and can then be used in our framework to generate data as well as associated labels. More interestingly, such labeling functions are often easily generalizable, thus allowing our framework to be extended to transfer learning and multi-task learning (discussed in Section \ref{sec_analysis}). Figure \ref{fig:Motivational_example} shows a few examples of our results to illustrate the overall idea.

\begin{figure*}
\centering
    \includegraphics[width=\textwidth, height=0.5\textwidth]{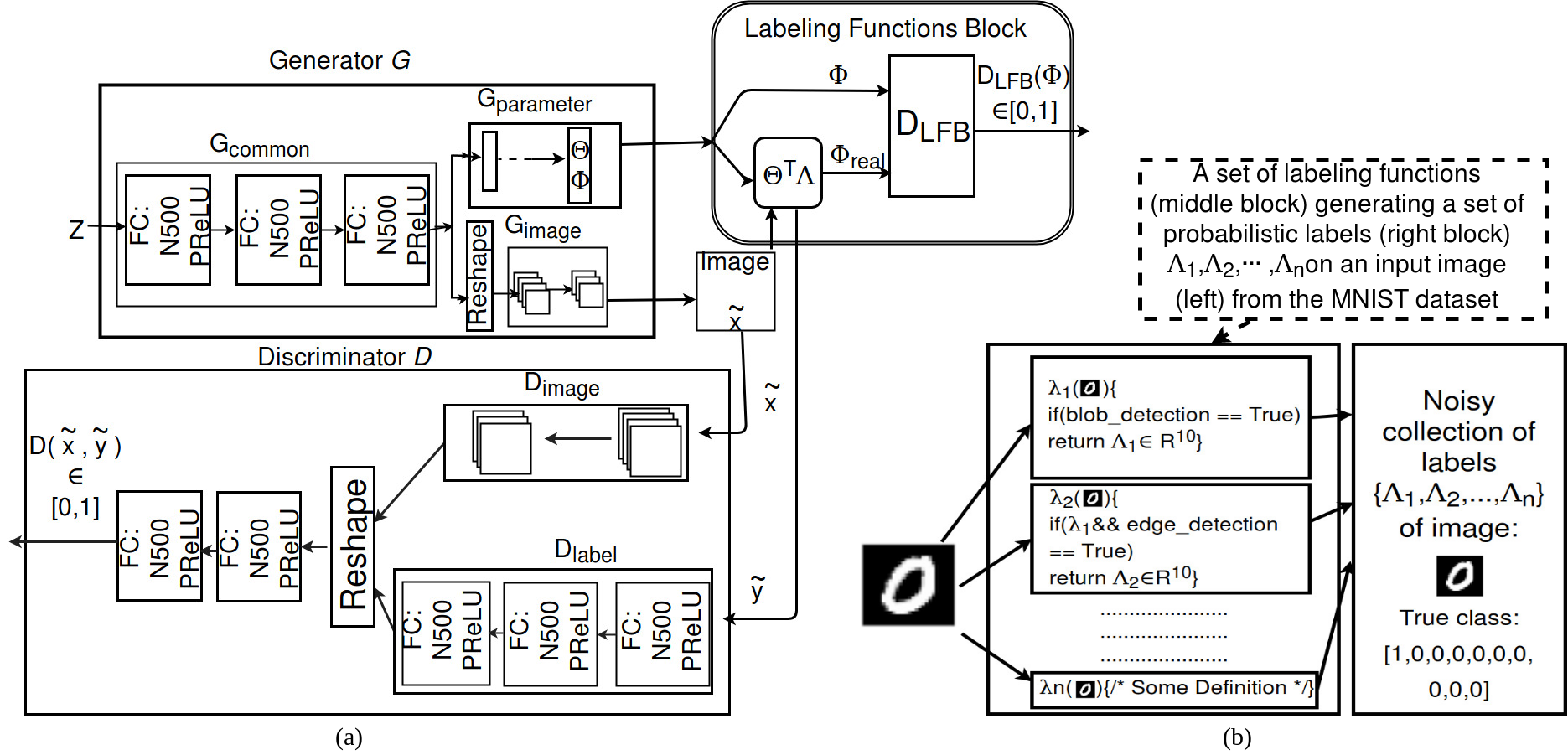}
    \caption{(a) Overall architecture of the Adversarial Data Programming (ADP) framework; (b) Example of a set of labeling functions}
    \label{fig:Diagram}
\end{figure*}

In practice, labeling functions can be associated with two kinds of dependencies: (i) relative accuracies, which measure the correctness of the labeling functions w.r.t. the true class label; and (ii) inter-function dependencies that capture the relationships between the labeling functions with respect to the predicted class label. In this work, we have proposed a novel adversarial framework using Generative Adversarial Networks (GANs) that learns these dependencies along with the data distribution using a minmax game. Our GAN learns to generate a joint data-label distribution using a generator block, a discriminator block and a Labeling Functions Block (LFB), which contains another discriminator that helps in learning the two kinds of dependencies mentioned above. The overall architecture of the proposed \model architecture is presented in Figure \ref{fig:Diagram}a.

Our broad idea of learning relative accuracies and inter-function dependencies of labeling functions is inspired by the recently proposed Data Programming (DP) framework \cite{ratner2016data} (and hence, the name ADP), but our method is different in many ways: (i) DP is a strict conditional model (i.e. $P(\tilde{y}|\mathbf{x})$) that requires additional unlabeled data points even at test time, while our model is a joint distribution model, i.e. $P(\mathbf{x},y)$, and does not require any additional unlabeled data points at test/generation time. (ii) DP learns a generative model using Maximum Likelihood Estimation (MLE) and gradient descent to learn the relative accuracies of labeling functions. We however replace this approach with a GAN-based adversarial estimation of parameters. \cite{danihelka2017comparison} and \cite{theis2015note} provide insights on the advantage of using a GAN-based estimator over MLE to achieve a relatively quicker training time and good robustness on generated samples. (iii) To learn the statistical dependencies of labeling functions, DP models the dependency structure of labeling functions as a factor graph, and uses computationally expensive Gibbs sampling techniques to update the gradient in each step. We replace the factor graph and Gibbs sampling-based estimation of inter-function dependencies with another discriminator in our GAN-based estimation, which speeds up the learning process again and provides a robust generation at run-time.

As our outcomes of this work, we show how a set of low-quality, weak labeling functions can be used within a framework that models a joint data-label distribution to generate robust samples. We also show that this idea can be generalized quite easily to transfer learning and multi-task learning settings, showing the generalizability of this work. Our contributions can be summarized as follows:
\begin{itemize}[noitemsep,topsep=0pt]
\item We propose a novel adversarial framework, \model to generate robust data-label pairs that be used to obtain datasets in domains that have very little data and thus save human labor and time.
\item We show how an adversarial framework can be used to learn dependencies between weak labeling functions and thus provide high-fidelity aggregated labels along with generated data in a GAN setting.
\item The proposed framework can also be used in a transfer learning setting where \model can be trained on a source domain, and then finetuned on a target domain to then generate data-label pairs in the target domain.
\item We also show the potential of this \model framework to generate cross-domain data in a multi-task setting, where images from two domains are generated simultaneously by the model along with the labels.
\end{itemize}
\section{Related Work}
Data augmentation seems a natural answer to the scarcity of curated hand-labeled training data. However, heuristic data augmentation techniques like \cite{dosovitskiy2016discriminative} and \cite{graham2014fractional} use a limited form of class-preserving image transformations such as rotation, mirroring, addition of small noise, random crop etc. Interpolation-based methods proposed in \cite{devries2017dataset} and class-conditional models of diffeomorphisms proposed in \cite{hauberg2016dreaming} interpolate between nearest-neighbor labeled data points. The popular SMOTE algorithm \cite{chawla2002smote} performs oversampling to reduce class imbalance and augment the given data. All of these methods vastly depends on hand-tuned parameters, the order of geometric transformations, the optimal value of transformation parameters, etc. A small change in parameters can often lead to negative impacts on final performance as studied in \cite{ratner2017learning}, \cite{ciresan3deep} and \cite{dosovitskiy2016discriminative}). 

In this work, we choose to use a more intuitive way of creating labeled data by learning a joint distribution model. Learning a joint data-label distribution using generative models such as \cite{doersch2016tutorial}, \cite{goodfellow2014generative}, and \cite{li2015generative} is non-trivial, since the label often requires domain knowledge and not directly inferrably from data. Our proposed model hence uses distant supervision signals (in the form of labeling functions) to \textit{generate} novel labeled data points. Distant supervision signals such as labeling functions are cheaper than manual annotation of each data point, and has been successfully used in recent methods such as \cite{ratner2016data}. Ratner et al. proposed a generative model in \cite{ratner2016data} that uses a fixed number of user-defined labeling functions to programatically generate synthetic labels for data in near-constant time. DP outperformed number of approaches such as multiple-instance learning (\cite{riedel2010modeling}), co-training (\cite{blum1998combining}), crowdsourcing (\cite{gao2011harnessing}), or ensemble based weak-learner method like boosting (\cite{schapire2012boosting}), thus reinforcing our choice in this work. Alfonseca et al. \cite{alfonseca2012pattern} generated additional training data using hierarchical topic models for weak supervision. Heuristics for distant supervision are also proposed in \cite{bunescu2007learning}, but this method does not model the inherent noise associated with such heuristics. Structure learning \cite{varma2017inferring}\cite{ratner2017learning} also exploits the use of distant supervision signals for generating labels, but as described in Section \ref{sec_intro}, these methods like \cite{ratner2016data} require unlabeled test data to generate a labeled dataset. Additionally, \cite{ratner2016data}, \cite{ratner2017learning} and \cite{varma2017inferring} are computationally expensive due to its use of Gibbs sampling in MLE. 

We instead use an adversarial approach to learn the joint distribution by weighting a set of domain-specific label functions using a Generative Adversarial Network (GAN). GAN (\cite{goodfellow2014generative}) approximates the real data distribution by optimizing a minmax objective function and thus generates novel out-of-sample data points. Broadly, GAN can be viewed in terms of three manifestations: (i) GANs can be trained to sample from a marginal distribution $P(\textbf{x})$ (\cite{denton2015deep}, \cite{radford2015unsupervised}\cite{arjovsky2017wasserstein}), where $\textbf{x}$ refers to data. (ii) Recent efforts in literature such as Conditional GAN \cite{mirza2014conditional}, Auxiliary Classifier GAN \cite{odena2016conditional} and InfoGAN \cite{chen2016infogan} show training of GANs conditioned on class labels $y$ to thus sample from a conditional distribution, i.e. $P(\textbf{x}|y)$. Other state-of-the-art models with similar objectives have exploited other modalities for the same purpose; for example, Zhang et al \cite{zhang2016stackgan} propose a GAN conditioned on images, while Hu et al \cite{hu2017controllable} propose a GAN conditioned on text. (iii) There have been a few very recent efforts \cite{yi2017dualgan}, \cite{zhu2017unpaired} and \cite{kim2017learning}), which attempt to train GANs to sample from a joint distribution. For example, CoGAN (\cite{liu2016coupled}) introduces a parameter-sharing approach to learn an unpaired joint distribution between two domains, while TripleGAN \cite{li2017triple} brings together a classifier along with the discriminator and generator which helps in a semi-supervised setting. In this work, we propose a novel idea to instead use distant supervision signals to accomplish learning the joint distribution of labeled images. We now describe the proposed methodology.
\section{Adversarial Data Programming (ADP): Methodology}
\label{sec_adp_methodology}
\noindent Our central aim in this work is to learn parameters of a probabilistic model:
\begin{equation}
\label{eq:joint_distribution}
P(\mathbf{x},y)    
\end{equation}
that captures the joint distribution over the data $\mathbf{x}$ and the corresponding labels $y$, thus allowing us to generate out-of-sample data points along with their corresponding labels (we focus on images in the rest of this paper).

While recent efforts such as \cite{liu2016coupled} and \cite{gan2017triangle} have considered complementary objectives, they largely focused on learning joint probability distributions in cross-domain understanding settings. In this work, we focus on learning the joint image-label probability distribution with a view to automatically create labeled datasets, by exploiting the use of distant supervision signals to generate labeled data. To the best of our knowledge, this is the first such work that 
invokes distant supervision while learning the joint distribution $P(\mathbf{x}, y)$, so as to generate labeled data points at scale from $P(\mathbf{x}, y)$. Besides, automatic generation of labels for data based on training data-label pairs is non-trivial, and often does not work directly. Distant supervision provides us a mechanism to achieve this challenging goal.  We encode distant supervision signals as a set of (weak) definitions by annotators using which unlabeled data points can be labeled. These definitions can be harvested from knowledge bases, domain heuristics, ontologies, rules-of-thumb, educated guesses, decisions of weak classifiers or obtained using crowdsourcing. Many application domains have such distant supervision available in different means through domain knowledge or heuristics, which can be leveraged in the proposed framework. We provide examples in Section \ref{sec_expts} when we describe our experiments. 

We encapsulate all available distant supervision signals, henceforth called \textit{labeling functions}, in a unified abstract container called \textbf{L}abeling \textbf{F}unctions \textbf{B}lock (LFB, see Figure \ref{fig:Diagram}a). Let LFB comprise of $n$ labeling functions $\lambda_1, \lambda_2,\cdots, \lambda_n$, where each labeling function is a mapping:
\begin{equation} 
\label{eq:labeling_functions}
\lambda_i: \mathbf{x}_j  \rightarrow \Lambda_{ij}
\end{equation}
that maps a data point $\mathbf{x}_j$ to a $m$-dimensional probabilistic label vector, $\Lambda_{ij} \in \mathbb{R}^m$, where $m$ is the number of class labels with $\sum_m \Lambda_{ij} = 1 \text{ and } 0 \leq \Lambda_{ij}^k \leq 1$ for each $k \in \{1,\cdots,m\}$. For example, $\mathbf{x}_j$ could be thought of as an image from the MNIST dataset, and $\Lambda_{ij} \in \mathbb{R}^{10}$ would be the corresponding label vector when the labeling function $\lambda_i$ is applied to $\mathbf{x}_j$. $\Lambda_{ij}$, for instance, could be the one-hot 10-dimensional class vector, see Figure \ref{fig:Diagram}b. 

We characterize the set of labeling functions, $\{\lambda_i, i=1,\cdots,n\}$, with two kinds of dependencies: (i) \textit{relative accuracies} of the labeling functions with respect to the true class label of a given data point; and (ii) \textit{inter-function dependencies} that capture the relationships between the labeling functions with respect to the predicted class label. To obtain a final label $y$ for a given data point $\mathbf{x}$ using the LFB, we use two different sets of parameters, $\Theta$ and $\Phi$ to capture each of these dependencies between the labeling functions. We, hence, denote the Labeling Function Block (LFB) as:
\begin{equation} 
\label{eq:labeling_functions_block}
LFB_{\lambda,\Theta,\Phi}: \mathbf{x}_j \rightarrow \Lambda_{j}
\end{equation}
i.e. given a set of labeling functions $\lambda$, a set of parameters capturing the relative accuracy-based dependencies between the labeling functions, $\Theta$, and a second set of parameters capturing inter-label dependencies, $\Phi$, $LFB$ provides a probabilistic label vector, $\Lambda_j$, for a given data input $\mathbf{x}_j$.

The joint distribution we seek to model in this work (Equation \ref{eq:joint_distribution}) hence becomes:
\begin{equation} 
\label{eq:modified_joint_distribution}
P(\mathbf{x}, LFB_{\lambda,\Theta,\Phi}(\mathbf{x}))
\end{equation}
In the rest of this section, we show how we can learn the parameters of the above distribution modeling image-label pairs using an adversarial framework with a high degree of label fidelity. We use Generative Adversarial Networks (GANs) to model the joint distribution in Equation \ref{eq:modified_joint_distribution}. In particular, we provide a mechanism to integrate the LFB (Equation \ref{eq:labeling_functions_block}) into the GAN framework, and show how $\Theta$ and $\Phi$ can be learned through the framework itself. Our adversarial loss function is given by:
\begin{equation} 
\label{eq:base_adversarial_function}
\begin{split}
\min \max L(G,D) = \mathbb{E}_{(\mathbf{x},y) \sim P_{real}(\mathbf{x},y)} \log(D(\mathbf{x}, y)) + \\
\mathbb{E}_{(\tilde{\mathbf{x}}, \Lambda) \sim P_{fake}(z)} \log(1-D(\tilde{\mathbf{x}}, \Lambda))
\end{split}
\end{equation}
where $G$ is the generator module and $D$ is the discriminator module. The overall architecture of the proposed ADP framework is shown in Figure \ref{fig:Diagram}a.

This approach has a few advantages: (i) firstly, labeling functions (which can even be just loosely defined) are cheaper to obtain than collecting labels for a large dataset; (ii) labeling functions can help bring domain knowledge into such generative models; (iii) labeling functions act as an implicit regularizer in the label space, thus allowing good generalization; (iv) with a small fine-tuning, labeling functions can be easily re-purposed for new domains (\textit{transfer learning}), as we describe later in this paper.

The ADP architecture is designed to learn the parameters required to model the joint distribution in Equation \ref{eq:modified_joint_distribution}, and thus generate out-of-sample image-label pairs. This architecture is broadly divided into three modules: the generator, discriminator and the LFB. We now describe each of these modules individually.
\subsection{The \model- Generator}
\vspace{-5pt}
Given a noise input $z$ and a set of labeling functions $\lambda$, the generator $G$ outputs an image $\mathbf{x}$ and the parameters $\Theta$ and $\Phi$, the dependencies between the labeling functions described earlier. In particular, $G$ consists of three blocks: $G_{common}$, $G_{image}$ and $G_{parameter}$, as shown in Figure \ref{fig:Diagram}a. $G_{common}$ captures the common high-level semantic relationships between the data and the label space, and is comprised only of fully connected (FC) layers. The output of $G_{common}$ forks into two branches: $G_{image}$ and $G_{parameter}$, where $G_{image}$ generates the image $\tilde{\mathbf{x}}$, and $G_{parameter}$ generates the parameters $(\Theta, \Phi)$. While $G_{parameter}$ uses FC layers, $G_{image}$ uses Fully Convolutional (FCONV) layers to generate the image (more details in Section \ref{sec_expts}). Thus, the generator $G$ outputs $(\mathbf{x}, \Theta, \Phi)$ given input $z \sim \mathcal{N}(0,I)$, the standard normal distribution.
\subsection{The \model- Discriminator}
\vspace{-5pt}
The discriminator $D$ of \model estimates the likelihood of an image-label input pair being drawn from the real distribution obtained from training data. $D$ takes a batch of image-label pairs as input and maps that to a probability score to estimate the aforementioned likelihood of the image-label pair. To accomplish this, $D$ has two branches: $D_{image}$ and $D_{label}$ (shown in the Discriminator block in Figure \ref{fig:Diagram}a). These two branches are not coupled in the initial layers, so as to separately extract required low-level features. The branches share weights in later layers to extract joint semantic features that help $D$ classify correctly if an image-label pair is \textit{fake} or \textit{real}.

We hence expand our objective function from Equation \ref{eq:base_adversarial_function} to the following:
\begin{equation} 
\label{eq:modified_adversarial_function}
\begin{split}
\min \max L(G,D_{image},D_{label}) = \qquad \qquad \qquad \qquad \qquad\\
\mathbb{E}_{(\mathbf{x},y) \sim P_{real}(\mathbf{x},y)} \log(D_{image}(\mathbf{x}))\\
+ \quad \mathbb{E}_{z \sim \mathcal{N}(0,I)} \log(1-D_{image}(G_{image}(z)))\\
+ \quad \mathbb{E}_{(\mathbf{x},y) \sim P_{real}(\mathbf{x},y)} \log(D_{label}(y))\\
+ \quad \mathbb{E}_{z \sim \mathcal{N}(0,I)} \log(1-D_{label}(LFB(G(z))))\\
\end{split}
\end{equation}
\subsection{The \model- Labeling Function Block}
\vspace{-5pt}
This is a critical module of the proposed ADP framework. Our initial work revealed that a simple weighted (linear or non-linear) sum of the labeling functions do not perform well in generating out-of-sample image-label pairs. We hence used a separate adversarial methodology within this block to learn the dependencies, both relative accuracies and inter-function (discussed earlier in this section), between the labeling functions provided to the framework. We describe the components of the LFB below.
\subsubsection{Relative Accuracies of Labeling Functions}
\vspace{-2pt}
\label{subsubsec_relativeaccuracies}
The output, $\Theta$, of the $G_{parameter}$ block in the ADP-Generator $G$ provides the relative accuracies of the labeling functions. Given the image output generated by $G_{image}$: $\tilde{\mathbf{x}}$, the labeling functions $\{\lambda_1,\cdots,\lambda_n\}$, and the probabilistic label vectors $\{\Lambda_{i},i=1,\cdots,n\}$ obtained using the labeling functions (as in Eqn \ref{eq:labeling_functions}), we define the aggregated final label as:
\begin{equation} 
\label{eq:learning_theta}
\tilde{y} = \sum_{i=1}^{n}\tilde{\theta}_{i} \Lambda_{i}\\
= \tilde{\Theta}\cdot\Lambda
\end{equation}
where $\tilde{\theta}_{i}$ is the normalized version of $\theta_{i}$, i.e. $\tilde{\theta}_{i} = \frac{\theta_{i}}{\sum_{k=1}^n \theta_{k}}$. The aggregated label, $\tilde{y}$, is provided as an output of the LFB.
\subsubsection{Inter-function Dependencies}
\vspace{-5pt}
Our empirical studies with considering only relative accuracies of labeling functions as a weighting mechanism led to \textit{mode collapse} in the joint distribution space, a well-understood problem in GANs. Our preliminary empirical studies demonstrated mode collapse in the joint distribution space (either images of same class with different labels, or images of different classes with same label were generated). The rationale behind taking two discriminators is to penalize the missing modes. Related literature [36] shows that inter-functional dependencies act as an implicit regularizer in the label space. We also conducted experiments on synthetic data to demonstrate this issue (please see Fig \ref{fig:synthetic_image} below).
\vspace{-5pt}
We hence introduced an adversarial mechanism inside the LFB to influence the final relative accuracies, $\tilde{\theta}$, using the inter-function dependencies between the labeling functions. $D_{LFB}$, a discriminator inside LFB, receives two inputs: $\Phi$, which is output by $G_{parameter}$, and $\Phi_{real}$, which is obtained from $\Theta$ using the procedure described in Algorithm \ref{alg:computing_real_phi}. 
\begin{algorithm}
\SetAlgoLined

\hspace*{\algorithmicindent} \textbf{Input:} Labeling functions $\{\lambda_1,\cdots,\lambda_n\}$, Relative accuracies $\theta_1,\cdots,\theta_n$, Output probability vectors of labeling functions $\Lambda_1,\cdots,\Lambda_n$\\
\hspace*{\algorithmicindent} \textbf{Output:} $\Phi_{real}$\\


Set $\Phi_{real} = \text{I}(n,n)$\; \tcc{\footnotesize{I = Identity Matrix}}
 
\For{$i = 1$ to $n$}{ \tcc{\footnotesize{For each labeling function}}

  \For{$j = i+1$ to $n$}{ \tcc{\footnotesize{For each other labeling function}}
  
     \tcc{\footnotesize{If one-hot encoding of the outputs of two functions match, increment $(i,j)$th entry in $\Phi_{real}$ by 1}}
     $\Phi_{real}(i,j) = \Phi_{real}(i,j) + \text{OneHot}(\theta_i \Lambda_i) \cdot \text{OneHot}(\theta_j \Lambda_j)$\;
  }
 }

\For{$p = 1$ to $n$}{
 $\Phi_{real}$(p, .) = $\frac{\Phi_{real}(p, .)}{\sum_{u=1}^n \Phi_{real}(p,u)}$\;
}

Set $\Phi_{real} = \Phi_{real} + \Phi_{real}^T - \text{diag}(\Phi_{real})$ \tcc{\footnotesize{Complete matrix using symmetry}}

\caption{Procedure to compute $\Phi_{real}$}
\label{alg:computing_real_phi}
\end{algorithm}

Algorithm \ref{alg:computing_real_phi} computes a matrix of interdependencies between the labeling functions, $\Phi_{real}$, by looking at the one-hot encodings of their predicted label vectors. If the one-hot encodings match for a given data input, we increase the count of their correlation by one, and compute this matrix across a particular mini-batch of data points under consideration. The counts are then normalized row-wise to obtain $\Phi_{real}$. The task of the discriminator is to recognize the computed interdependencies as real, and the $\Phi$ generated through the network in $G_{parameter}$ as fake. The gradient backpropagated through this discriminator to the $G$ block is critical as a regularizer in learning a better $\Theta$, which is finally used to weight the labeling functions (as in Section \ref{subsubsec_relativeaccuracies}). Combining the gradient information from $D_{LFB}$ along with $D$, penalizes missing modes and helps $G$ to generate more variety in the samples. The objective function of our second adversarial module is hence:
\begin{equation} 
\label{eq:second_adversarial_function}
\begin{split}
\min \max L(D_{LFB}, G) = \qquad \qquad \\
+ \quad \mathbb{E}_{z \sim \mathcal{N}(0,I)} \log(D_{LFB}(\Phi_{real}(z)))\\
+ \quad \mathbb{E}_{z \sim \mathcal{N}(0,I)} \log(1-D_{LFB}(\Phi(z)))
\end{split}
\end{equation}
\noindent where $\Phi_{real}$ and $\Phi$ are obtained from $G_{parameter}(z)$ as described above. More details of the LFB are provided in implementation details in Section \ref{sec_expts}.

The overall architecture of \model(Figure \ref{fig:Diagram}a) is trained using end-to-end backpropagation with gradients from both discriminators, $D$ and $D_{LFB}$, influencing the weights learned inside the generator $G$. Mini-batches of image-label pairs from a given training distribution are provided as input to \model\xspace, and Stochastic Gradient Descent (SGD) is used to learn the parameters of the model. At the end of training, we define the aggregated final label as:
\begin{equation} 
\label{eq:learning_phi}
\tilde{y} = \tilde{\Theta}\cdot \Phi^{T}\cdot\Lambda
\end{equation}the samples $(\tilde{\mathbf{x}},\tilde{y})$ generated using the $G$ and $LFB$ modules thus provide samples from the desired joint distribution (Eqn \ref{eq:joint_distribution}) modeled using the framework.
\section{Experiments and Results}
\label{sec_expts}
\subsection{Datasets}
\vspace{-5pt}
We validated the ADP framework on standard datasets: MNIST (\cite{MNIST}), Fashion MNIST (\cite{Fashion-MNIST}), SVHN (\cite{SVHN}), and CIFAR-10 (\cite{CIFAR10}). No additional pre-processing is performed on MNIST, Fashion-MNIST and CIFAR 10 datasets. For the SVHN dataset, we used the `Format 2 Cropped version', and included an additional crop on each image to reduce presence of more than one digit, though the dimension $32 \times 32$ is maintained\footnote{Code available at \url{https://github.com/ArghyaPal/Adversarial-Data-Programming}}. 
\subsection{Labeling Functions}
\vspace{-5pt}
Labeling functions form a critical element of \model\xspace, and we used different cues from state-of-the-art algorithms to help obtain labeling functions for our experiments. Table \ref{table:cues_for_MNIST_SVHN} shows the labeling functions we used for our experiments on MNIST and SVHN (digit recognition problems), and Table \ref{table:cues_for_CIFAR_FMNIST} shows the functions used for CIFAR and Fashion-MNIST. We categorized labeling functions as: (i) Heuristic; (ii) Image processing-based; and (iii) Deep learning-based labeling functions (as in Tables \ref{table:cues_for_MNIST_SVHN} and \ref{table:cues_for_CIFAR_FMNIST}). Table \ref{table:labeling_function_stat} presents the statistics of the number of labeling functions used for each of the considered datasets (the empirical study that motivated these choices is presented in Section \ref{sec_analysis}). In this work, for each labeling function, a simple threshold rule on the $L_2$-norm of the aforementioned features is used, where the threshold is obtained empirically as the mean of the  $L_2$-norms of a randomly chosen subset, which is $\alpha$-trimmed to remove outliers. 
More examples of labeling functions and ablation studies on their usefulness are presented in the Supplementary Section.
\begin{table}[h]
\small
\centering
\begin{tabular}{|p{2cm}|p{5.5cm}|}
\hline
\textbf{Type} & \textbf{Labeling Functions used} \\ 
\hline
Heuristic & Presence of long edges (vertical or horizontal) \cite{mandal2011handwritten}; Image histogram\\
\hline
Image Processing based & Bag-of-feature \cite{rothacker2012bag}; Haar wavelet \cite{chen2014unsupervised}; Discrete-continuous ADM \cite{laude2017discrete}; Compressive sensing \cite{zhenjiang1994handwritten}\\
\hline
Deep Learning based & Convolution kernels from last conv layer (before fully connected layers) of LeNet\\
\hline
\end{tabular}
\caption{Labeling functions used for MNIST and SVHN datasets, both of which represent the digit recognition task}
\label{table:cues_for_MNIST_SVHN}
\end{table}

\begin{table}[h]
\small
\centering
\begin{tabular}{|p{2cm}|p{5.5cm}|}
\hline
\textbf{Type} & \textbf{Labeling Functions used} \\ 
\hline
Heuristic & PatchMatch \cite{barnes2011patchmatch}; Blob Detection; Presence of edges \cite{alfonseca2012pattern}; Textons; Image histogram\\
\hline
Image Processing based & Global descriptor (GIST-based) \cite{moudni2013recognition}; Local descriptor (SIFT-based) \cite{yu2011learning}; Bag-of-visual-words; Histogram of Oriented Gradient (HOG)-based: HoGgles \cite{vondrick2013hoggles}\\
\hline
Deep Learning based & Convolution kernels from last conv layer (before fully connected layers) of (ImageNet) pre-trained AlexNet\\
\hline
\end{tabular}
\caption{Labeling functions used for CIFAR 10 and Fashion MNIST datasets}
\label{table:cues_for_CIFAR_FMNIST}
\end{table}

\begin{table}
\centering
\scalebox{0.8}{
    \begin{tabular}{|c|c|c|c|}
        \hline
         & Heuristic & Image Processing & Deep Learning \\ \hline
        MNIST & 43 & 10 & 1 \\ \hline
        Fashion-MNIST & 50 & 6 & 1 \\ \hline
        SVHN & 43 & 10 & 2 \\ \hline
        CIFAR 10 & 46 & 18 & 2 \\ \hline
    \end{tabular}
}
\caption{Number of labeling functions used for different datasets}
\label{table:labeling_function_stat}
\end{table}
\begin{figure*}[t!]
\centering
\includegraphics[width=0.97\textwidth,height=0.3\textwidth]{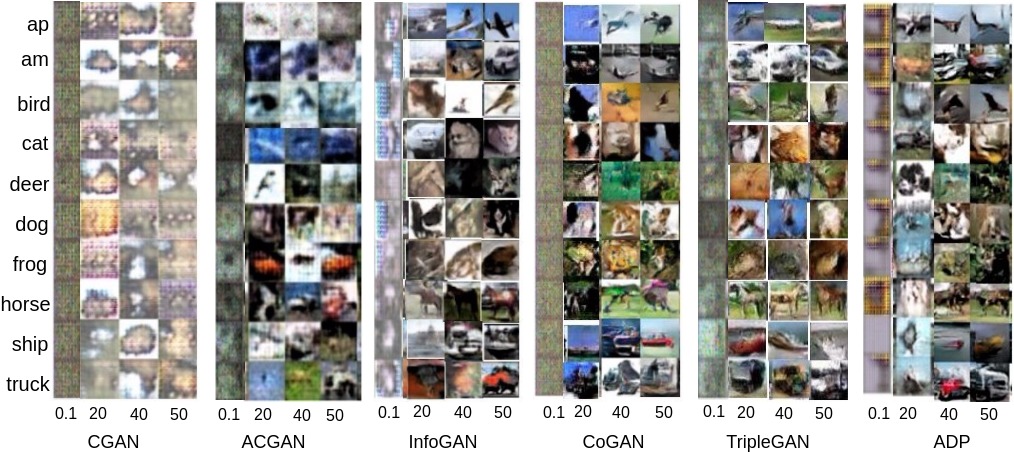}
\caption{(Best viewed in color) Image-label pairs generated by training on CIFAR10 dataset using CGAN, ACGAN, InfoGAN, CoGAN, TripleGAN and our method, \model. For a given model, the columns of images represents generations after 0.1k, 20k, 40k, 50k epochs, and the rows correspond to the associated class label. `ap' stands for airplane, and `am' stands for the automobile class of CIFAR 10 dataset. Note the clarity of generations of the proposed method.}
\label{fig:Qualitative_comparison}
\end{figure*}
\subsection{Implementation Details}
\vspace{-5pt}
$G_{common}$ has 3 dense fully connected layers (FC) (128 nodes per layer) with batch-normalization. $G_{image}$ continues with fractional length convolutional layers, similar to \cite{liu2016coupled} (FCONV: 1024 nodes per layer, Kernel size: $4 \times 4$, Stride: 1, followed by batch-normalization and Parameterized ReLU), and generates image $\textbf{x}$. $G_{parameter}$ uses FC layers and generates $\Theta, \Phi$. The discriminator network $D$ follows the \say{in-plane rotation} network of \cite{liu2016coupled}. $D_{label}$ is a stack of FC layers. Both $D_{image}$ and $D_{LFB}$ have 2 FCONV layers followed by FC layers. We trained the complete model with mini-batch Stochastic Gradient Descent (SGD) with a batch size of 128, learning rate of 0.0001, momentum factor of 0.5 and Adam as an optimizer. 
\subsection{Comparison with State-of-the-Art Models}
\label{subsec_compare}
\begin{figure}
    \centering
    \includegraphics[width=0.45\textwidth]{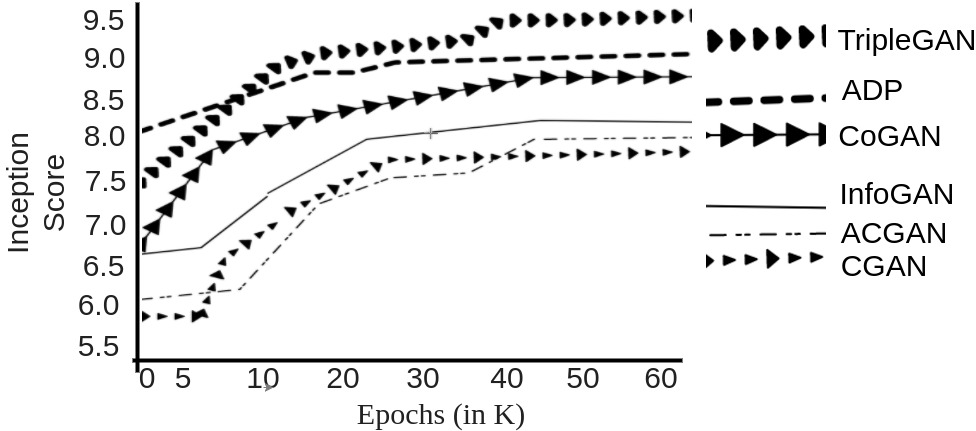}
    \caption{Inception scores on CIFAR 10 (Section \ref{subsec_compare}, Quantitative Analysis)}
    \label{fig:Inception_Score}
\end{figure}

\vspace{-5pt}
\paragraph{Qualitative Results:} We compared our method against other generative methods that allow generation of data along with a label: Conditional GAN or CGAN (\cite{goodfellow2014generative}), ACGAN (\cite{odena2016conditional}), InfoGAN (\cite{chen2016infogan}), CoGAN (\cite{liu2016coupled}) and TripleGAN (\cite{li2017triple}). (We changed the use case setup of these methods to generate data-label pairs as required. For example, for a conditional GAN, we specified a class label, generated a corresponding image and used this as an image-label pair.) We used the publicly available codes for each of the above methods, and the results for CIFAR10 are shown in Figure \ref{fig:Qualitative_comparison} (Results for other datasets are shown in the Supplementary Section). The figure shows that the proposed model generates images with very good clarity. Besides, while some of the aforementioned methods (such as CGAN and InfoGAN) generate images conditioned on a given label (and hence require a label to be provided as input), the label is provided by the model in our case.
\paragraph{Quantitative Results:} We considered three evaluation metrics for studying the performance of our method quantitatively: (i) \textit{Human Turing test (HTT):} This metric studies how hard it is for a human annotator to tell the difference between real and generated samples. We asked 40 subjects to evaluate  image quality and image-label correspondence (Table \ref{table:HTT_Results}) on a scale of 10, given 50 random image-label samples from the generated pool for each method considered. Table \ref{table:HTT_Results} shows consistently good performance of \model over other methods, especially in image-label correspondence, which is the focus of this work; (ii) \textit{Inception Score:} The inception score, as used in \cite{liu2016coupled} and \cite{li2017triple}, for the CIFAR 10 dataset is shown in Figure \ref{fig:Inception_Score}. The figure shows that \model and TripleGAN perform significantly better than the rest of the methods (more results on other datasets included in the Supplementary Section). We also used a \textit{Parzen window based evaluation} metric, and these results are included in the Supplementary Section. 

We trained a ResNet-56 model on the CIFAR-10 dataset under different settings, and the results are shown in Table \ref{table:ResNet}. The addition of labeled data generated using our method significantly lowers the test cross-entropy loss across the epochs. We will include these in the paper.

\vspace{-10pt}
\begin{table}[h]
    \centering
    \scalebox{0.7}{
        \begin{tabular}{|p{4.5cm}|c|c|c|c|c|c|c|}
            \hline \hline
            (Training Data, Test Data)&\multicolumn{7}{|c|}{Epochs}\\ \hline
            & 5k & 10k & 15k & 20k & 30k & 40k & 50k \\ \hline \hline
            (Real data-50K, Real data-10K) & 9.83 & 7.3 & 7.12 & 6.3 & 6.1 & 4.3 & 4.19\\ \hline
            (ADP data-50K, Real data-10K) & 9.32 & 8.9 & 8.13 & 7.0 & 6.75 & 5.53 & 5.0\\ \hline
            (Real data-50K, ADP data-10K) & 9.67 & 9.4 & 7.92 & 7.3 & 6.81 & 6.18 & 5.6\\ \hline
            (ADP-25K + Real data-25K, Real data-10K) & 8.5 & 6.6 & 6.21 & 5.7 & 5.5 & 4.83 & 3.5\\ \hline
            (\textbf{ADP-50K + Real data-50K, Real data-10K}) & \textbf{7.71} & \textbf{6.3} & \textbf{6.0} & \textbf{5.34} & \textbf{3.1} & \textbf{2.92} & \textbf{2.71}\\ \hline \hline
        \end{tabular}
    }
    \caption{Test cross-entropy loss of ResNet-56 on CIFAR-10 dataset. (Real data-50K, Real data-10K) = standard dataset; ADP = our method; 10/25/50K = the number of data points used in thousands. In ADP-25K + Real data-25K, class ratios were maintained as in the original dataset.}
    \label{table:ResNet}
    \end{table}
\begin{table*}
\footnotesize
\centering
\begin{tabular}{||p{0.9cm}||p{0.07\linewidth}|p{0.07\linewidth}|p{0.07\linewidth}|p{0.07\linewidth}|p{0.07\linewidth}||p{0.07\linewidth}|p{0.07\linewidth}|p{0.07\linewidth}|p{0.07\linewidth}|p{0.07\linewidth}||}
\hline \hline
\textbf{Dataset} & \multicolumn{5}{|c||}{\textbf{Image Quality}} & \multicolumn{5}{|c||}{\textbf{Image-Label Correspondence}}\\
\hline \hline
& ACGAN & CGAN & InfoGAN & TripleGAN & ADP & ACGAN & CGAN & InfoGAN & TripleGAN & ADP\\
\hline \hline
MNIST & $9.02 \pm 0.1$ & $9.11 \pm 0.2$ & $9.54 \pm 0.3$ & $\mathbf{9.6 \pm 0.4}$ & $9.46 \pm 0.3$ & $8.27 \pm 0.3$ & $9.11 \pm 0.2$ & $9.78 \pm 0.2$ & $9.6 \pm 0.4$ & $\mathbf{9.92 \pm 0.1}$\\
\hline
FMNIST & $9.32 \pm 0.4$ & $8.89 \pm 0.3$ & $9.10 \pm 0.3$ & $9.2 \pm 0.2$ & $\mathbf{9.33 \pm 0.6}$ & $8.8 \pm 0.1$ & $8.89 \pm 0.3$ & $9.27 \pm 0.4$ & $9.2 \pm 0.2$ & $\mathbf{9.93 \pm 0.1}$ \\
\hline
SVHN & $5.3 \pm 0.2$ & $4.91\pm 0.5$ & $7.71 \pm 0.1$ & $8.6 \pm 0.3$ & $\mathbf{8.86 \pm 0.3}$ & $8.53 \pm 0.3$ & $8.91\pm 0.0$ & $9.08 \pm 0.1$ & $\mathbf{9.75 \pm 0.2}$ & $9.72 \pm 0.3$ \\
\hline
CIFAR10 & $4.17 \pm 0.1$ & $4.36 \pm 0.2$ & $6.23 \pm 0.2$ & $\mathbf{8.5 \pm 0.1}$ & $8.27 \pm 0.3$ & $7.27 \pm 0.1$ & $8.62 \pm 0.2$ & $\mathbf{9.72 \pm 0.1}$ & $9.68 \pm 0.3$ & $9.49 \pm 0.5$\\
\hline \hline
\end{tabular}
\caption{Human Turing Test for image quality and image-label correspondence (Section \ref{subsec_compare}, higher the better). Note that the proposed method, \model performs the best in most cases, and is a close second when TripleGAN wins.}
\label{table:HTT_Results}
\end{table*}

\begin{figure}
    \centering
    \includegraphics[width=0.45\textwidth]{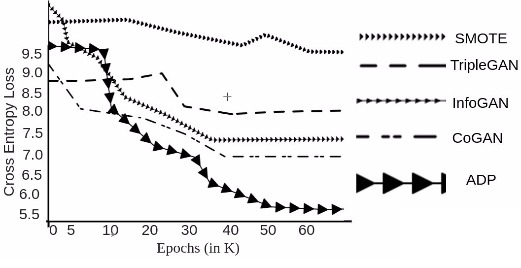}
    \caption{Classification performance of a pretrained ResNet model on image-label pairs generated by various models trained on CIFAR 10}
    \label{fig:classification_error}
\end{figure}

\paragraph{Classification Performance:} To study the usefulness of the generated image-label pairs, we studied the classification cross-entropy loss of a pretrained ResNet model on the image-label pairs generated by our \model at test time. A lesser cross-entropy loss in ResNet at test time indicates the efficacy of our model as a data augmentation method. We compared our method against TripleGAN, InfoGAN, CoGAN as well as the popular oversampling technique, SMOTE \cite{chawla2002smote}. Figure \ref{fig:classification_error} shows the result, which shows that the proposed model has significantly lower cross-entropy loss over other methods, highlighting its usefulness.
\section{Discussion and Analysis}
\label{sec_analysis}
\vspace{-5pt}
\paragraph{Optimal Number of Labeling Functions:} We studied the performance of \model when the number of labeling functions is varied to understand the impact of this parameter on the performance. We studied the test cross-entropy error of a pretrained ResNet model with image-label pairs generated by ADP, trained using different number of labeling functions. Table \ref{table:optimum_number_of_labeling_functions} shows our results, suggesting that 50-55 labeling functions provides the best performance, depending on the dataset. This justifies our choice of number of labeling functions in Table \ref{table:labeling_function_stat}.
\begin{table}
\footnotesize
\centering
\begin{tabular}{|p{2cm}|l|l|l|l|}
\hline
No of Labeling Functions & MNIST & F-MNIST & CIFAR10 & SVHN\\
\hline
3 & 70.23\% & 81.02\% & 87.39\% & 83.82\% \\ \hline
10 & 47.92\% & 71.52\% & 60.11\% & 75.91\% \\ \hline
\textit{25} & \textit{20.32\%} & \textit{30.53\%} & \textit{42.31\%} & \textit{38.30\%} \\ \hline
30 & 4.56\% & 12.47\% & 21.19\% & 26.66\% \\ \hline
40 & 1.40\% & 6.81\% & 19.93\% & 16.62\% \\ \hline
50 & \textbf{1.33\%} & 4.92\% & 18.93\% & 13.05\% \\ \hline
55 & 1.34\% & \textbf{4.80\%} & \textbf{18.45}\% & \textbf{12.83}\% \\ \hline
65 & 1.31\% & 4.73\% & 18.43\% & 12.82\% \\ \hline
70 & 1.25\% & 4.75\% & 18.40\% & 12.80\% \\ \hline
\end{tabular}
\caption{Performance of \model when number of labeling functions is varied (Section \ref{sec_analysis}, Optimal Number of Labeling Functions).}
\label{table:optimum_number_of_labeling_functions}
\end{table}
\paragraph{Transfer Learning:} The use of distant supervision signals such as labeling functions (which can often be generic) allows us to extend the proposed \model model to a transfer learning setting. In this setup, we trained \model initially on a source dataset and then finetuned the model to a target dataset, with very limited training. In particular, we first trained \model on the MNIST dataset, and subsequently finetuned the $G_{image}$ branch alone with the SVHN dataset. We note that the weights of $G_{common}$, $G_{parameter}$ and $D_{LFB}$ are unaltered. The final finetuned model is then used to generate image-label pairs (which we hypothesize will look similar to SVHN). Figure \ref{fig:Motivational_example} shows encouraging results of our experiments in this regard. 
\begin{figure}
    \centering
    \includegraphics[width=0.48\textwidth,height=0.05\textwidth]{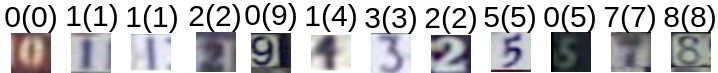}
    \caption{Transfer learning from MNIST to SVHN dataset. Digits within parentheses indicate true label, while the other is the label generated using our method (Section \ref{sec_analysis}, Transfer Learning)}
    \label{fig:GEN_CIFAR10}
\end{figure}

\begin{figure}
    \centering
    \includegraphics[height=0.23\textwidth, width=0.46\textwidth]{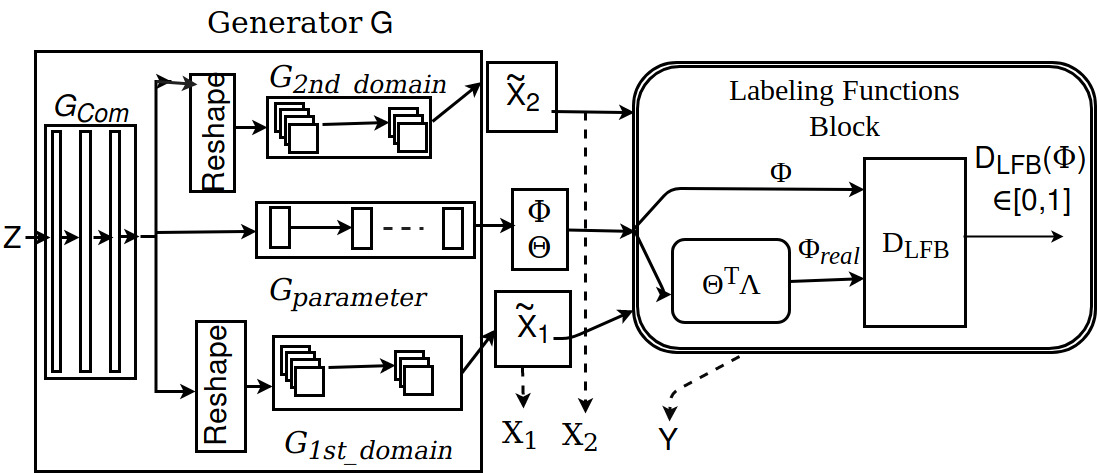}
    \caption{\model for Multi-Task Learning: Proposed Architecture}
    \label{fig:cross_domain_model}
\end{figure}

\begin{figure}
    \centering
    \includegraphics[height=0.25\textwidth, width=0.35\textwidth]{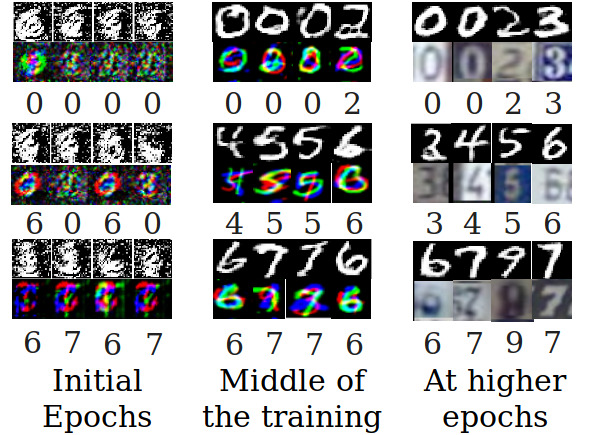}
    \caption{Results of \model- Multi-Task Learning on MNIST (black and white) and SVHN (RGB) datasets}
    \label{fig:cross_domain_1}
\end{figure}

\paragraph{Multi-task Joint Distribution Learning:} Learning a cross-domain joint distribution from heterogeneous domains is a challenging task. We show that the proposed \model method can be used to achieve this, by modifying its architecture as shown in Figure \ref{fig:cross_domain_model}, to simultaneously generate data from two different domains. We study this architecture on the MNIST and SVHN datasets, and show the promising results of our experiments in Figure \ref{fig:cross_domain_1}. The LFB acts as a regularizer and maintains the correlations between the domains in this case. More results on other datasets - in particular, LookBook and Fashion MNIST - are included in the Supplementary Section as well as Figure \ref{fig:Motivational_example}.
\section{Conclusions}
\vspace{-5pt}
Paucity of large curated hand-labeled training data for every domain-of-interest forms a major bottleneck in deploying machine learning methods in practice. Standard data augmentation techniques and other heuristics are often limited in their scope and require carefully picked hand-tuned parameters. We instead propose a new adversarial framework called Adversarial Data Programming (ADP), which can learn the joint data-label distribution effectively using a set of weakly defined labeling functions. The method shows promise on standard datasets, as well as on settings such as transfer learning and multi-task learning. Our future work will involve understanding the theoretical implications of this new framework from a game-theoretic perspective, as well as explore the performance of the method on more complex datasets.

{\small
\bibliographystyle{ieee}
\bibliography{egpaper_final}
}
\clearpage

\begin{center}
\large{\textbf{Supplementary Section}}
\end{center}
In this section, we include more details about our algorithm, labeling functions, datasets as well as additional results and comparisons, which could not be included in the main paper due to space constraints. 
\appendix
\section{Algorithm}
Algorithm \ref{alg:main} presents the overall stepwise routine of the proposed method, \model, as described in Section \ref{sec_adp_methodology}. During the training phase, the algorithm updates weights of the model by estimating gradients for a batch of labeled data points. The hyperparameters that need to be provided include standard parameters that are provided while training a GAN, such as: (i) number of iterations of Algorithm \ref{alg:main}; (ii) parameter $k$ (similar to \cite{goodfellow2014generative}) that describes how many times $D$ and $D_{LFB}$ would be updated with respect to $G$; and (iii) minibatch size $m$. Using empirical studies, we chose $m=128$, $k=2$ and number of iterations to be $60,000$.
\begin{algorithm*}
\SetAlgoLined
\hspace*{\algorithmicindent} \textbf{Input:} Number of iterations, Number of steps to train $D$: $k$, Minibatch size: $m$\\
\hspace*{\algorithmicindent} \textbf{Output:} Trained ADP model\\
\For{number of iterations}{
    \For{$k$ steps}{
        Given noise prior $\textbf{z} \sim \mathcal{N}(0,I)$, draw a batch of $m$ samples from $G$: $\{(\tilde{\mathbf{x}}_1, \Theta_1, \Phi_1),\cdots, (\tilde{\mathbf{x}}_m, \Theta_m, \Phi_m)\}$ \;
        Use Equation \ref{eq:labeling_functions_block} (from $LFB$) to compute probabilistic label vectors $\{\Lambda_1,\cdots,\Lambda_m\}$ given $\{(\tilde{\mathbf{x}}_1, \Theta_1, \Phi_1),\cdots, (\tilde{\mathbf{x}}_m, \Theta_m, \Phi_m)\}$\;
        Draw a batch of $m$ image-label pairs $((\textbf{x}_1, y_1),\cdots, (\textbf{x}_m, y_m))$ from real distribution $P_{real}(\textbf{x}, y)$\;
        Update weights of discriminators $D$ and $D_{LFB}$ ($\psi_{d}$ and $\psi_{l}$ respectively), using mini-batch stochastic gradient ascent with gradients as computed below:
        \[\nabla_{\psi_{d}} \frac{1}{m}\sum_{i=1}^{m}\Big[\log D(\textbf{x}_i, y_i) + \log (1- D(\tilde{\mathbf{x}}_i, \Lambda_i))\Big]\]and,
        \[\nabla_{\psi_{l}} \frac{1}{m}\sum_{i=1}^{m}\Big[\log D_{LFB}(\Phi_{real_i}) + \log (1 - D_{LFB}(\Phi_i))\Big]\]
        }
    Given noise prior $\textbf{z} \sim \mathcal{N}(0,I)$, draw a batch of $m$ samples from $G$: $\{(\tilde{\mathbf{x}}_1, \Theta_1, \Phi_1),\cdots, (\tilde{\mathbf{x}}_m, \Theta_m, \Phi_m)\}$\;
    Update weights of generator $G$, $\psi_{g}$, using mini-batch stochastic gradient descent with gradients as computed below (each step below updated sequentially, one after another)\;
        \[\nabla_{\psi_{g}} \frac{1}{m}\sum_{i=1}^{m}\Big[\log (1 - D_{LFB}(\Phi_i))\Big]\] and
        \[\nabla_{\psi_{g}} \frac{1}{m}\sum_{i=1}^{m}\Big[\log (1- D(\tilde{\mathbf{x}}_i, \Lambda_i))\Big]\]
}
\caption{Training \model}
\label{alg:main}
\end{algorithm*}
\section{Datasets}
In this section, we provide more information on the datasets used for validating \model in this work: MNIST, Fashion MNIST, SVHN and CIFAR 10. The MNIST dataset comprises $28 \times 28$ grayscale images (with one handwritten digit in each image) along with the corresponding label, with 50,000 training samples (image-label pairs). In case of SVHN, we used \say{format 2} of the dataset, which comprises 73257 $32 \times 32$ images (each containing a digit captured from street views of house numbers) with the corresponding labels. In case of CIFAR 10, we merged five training batches of the dataset and built a training set of 50,000 images. This dataset contains RGB-images each of size of $32 \times 32$ spanning 10 classes: \textit{automobile, airplane, bird, cat, deer, dog, frog, horse, ship, truck}. The total number of samples are almost equally distributed across all classes. Fashion MNIST, similar to MNIST, consists of a training set of 50,000 $28 \times 28$ grayscale images with one of 10 classes: \textit{Tshirt, Trouser, Pullover, Dress, Coat, Sandal, Shirt, Sneaker, Bag, Ankle boot}. 

\begin{figure}
\centering
\includegraphics[width=0.45\textwidth, height=0.5\textwidth]{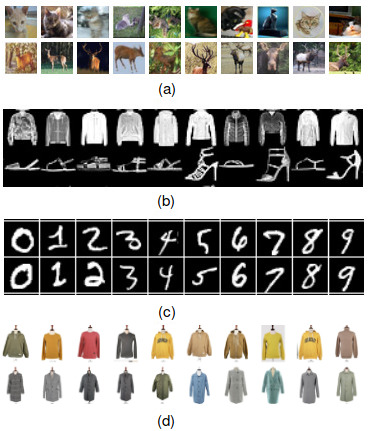}
\caption{Sample images from the datasets studied in this work: (a) CIFAR 10, (b) Fashion MNIST, (c) MNIST,  (d) LookBook}
\label{fig:Samples_of_real_dataset}
\end{figure}

We also used the LookBook \cite{yoo2016pixel} dataset (Figure \ref{fig:Samples_of_real_dataset}) to demonstrate cross-domain multi-task learning using \model, as described in Section \ref{sec_analysis}. This dataset contains 84,748 images across 17 classes: \textit{Midi dress, mini dress, coat, jacket, fur jacket, padded jacket, hooded jacket, jumper, cardigan, knitwear, blouse, shirt, sleeveless tee, short sleeve tee, long sleeve tee, hoody, vest}. In this work, we grouped these 17 classes into 4 classes: \textit{coat, pullover, t-shirt, dress}, in order to match with the Fashion MNIST dataset and thus help study cross-domain learning. We grouped coat, jacket, fur jacket, padded jacket, hooded jacket, jumper, cardigan to a single \textit{coat} class; hoody to the \textit{pullover} class; sleeveless tee, short sleeve tee to the \textit{t-shirt} class; cardigan, knitwear, blouse, Midi dress, mini dress to the \textit{Dress} class. Fashion MNIST dataset also has the same classes: \textit{coat, pullover, t-shirt, Dress} among its label, thus facilitating our study.

No additional pre-processing was performed on MNIST, Fashion MNIST, CIFAR 10 or the LookBook datasets. In case of SVHN, an additional crop was performed on each image to ensure only one digit is present in the image. The cropped image was subsequently sampled to maintain the $32 \times 32$ size.  Figure \ref{fig:Samples_of_real_dataset} shows illustrative examples of images from the chosen datasets.
\section{More on Labeling Functions}
The Labeling Functions Block (LFB) in Figure \ref{fig:Diagram}a (Section \ref{sec_adp_methodology}) is implemented using the open-source framework, Snorkel \cite{ratner2016data}. We modified the underlying architecture of the Snorkel framework to include an adversarial approach, which otherwise estimates dependencies using MLE invoking Gibbs sampling. Labeling functions of three kinds: heuristics, image processing-based and deep learned features have been used in this work, as described in Section \ref{sec_expts}. Examples of labeling functions used in this work are shown as Labeling Functions \ref{alg:HLF1}, \ref{alg:HLF2}, \ref{alg:HLF3} and \ref{alg:HLF4}. For each labeling function, a simple threshold rule on the $L_2$-norm of the aforementioned features is used. For each class of a dataset, the threshold information is obtained empirically as the average $L_2$-norm of the feature vectors of 20 random samples of that class (with $\alpha$-trimming to remove outliers). It is worthy to mention that, for an abstract understanding of working process of our labeling functions, the return value of example Labeling Functions \ref{alg:HLF3} and \ref{alg:HLF4} are one-hot encoding. Though in practice we fit a nonlinear function to get a probabilistic output.

\setcounter{algocf}{0}

\begin{algorithm}
\SetAlgorithmName{Labeling Function}{}{}
\SetAlgoLined
\hspace*{\algorithmicindent} \textbf{Input:} Image\\
\hspace*{\algorithmicindent} \textbf{Output:} Probabilistic label vector\\
\tcc{\footnotesize{Decision tree for English numerals recognition \cite{kumar2010classification}}}
\If{blob(Image) == TRUE}{
		\If{blob diameter(Image) $\leq$ 0.5cm}{
    	number = count blob(Image)\;
        \If{number == 2}{
        	return [0.2,0,0,0,0,0.1,0,0,0.6,0.1]\;
            }
        \If{number == 1}{return [0.6, 0, 0.2, 0, 0.1, 0, 0, 0, 0.1]}
        }
        \If{blob diameter(Image) $>$ 0.5cm}{return [0.4, 0, 0, 0, 0.1, 0, 0.3, 0.1, 0.1]}
}
\caption{Sample heuristic labeling function (used for blobs in digits like: 0, 9, 6)}
\label{alg:HLF1}
\end{algorithm}

\begin{algorithm}
\SetAlgorithmName{Labeling Function}{}{}
\SetAlgoLined
\hspace*{\algorithmicindent} \textbf{Input:} Image\\
\hspace*{\algorithmicindent} \textbf{Output:} Probabilistic label vector\\
\tcc{\footnotesize{Decision tree for English numerals recognition \cite{kumar2010classification}}}
\If{blob(Image) == TRUE}{
        number = count stem (Image)\;
        \If{number == 0}{return [0.8, 0, 0, 0, 0.1, 0, 0, 0, 0.1]}
        \If{number == 1}{return [0.1, 0, 0, 0, 0.4, 0, 0.4, 0, 0.1]}
        \If{number == 2}{return [0, 0, 0, 0, 0.8, 0, 0, 0, 0.2]}
    }
\caption{Sample heuristic labeling function (used for digits with blob and stem like: 4, 6, 9)}
\label{alg:HLF2}
\end{algorithm}

\begin{algorithm}
\SetAlgorithmName{Labeling Function}{}{}
\SetAlgoLined
\hspace*{\algorithmicindent} \textbf{Input:} Image, $n$: Number of classes\\
\hspace*{\algorithmicindent} \textbf{Output:} Probabilistic label vector\\
\tcc{\footnotesize{NOTE: Loop presented below for purposes of clarity - it is implemented only once for a dataset}}
\For{i=1$\cdots$ n}{
$v_{avg_i}$ = average value of $L_{2}$ norm of Bag-of-feature() on subset of training samples from class $i$\;
}
$\textbf{v} = Bag-of-feature (Image)$\;
$v_{image}$ = $\|\textbf{v}\|$\;
return OneHot($\argminA_i \Big[|v_{avg_i} - v_{Image}|\Big]$)
\caption{Sample image processing based labeling function (based on Bag-of-Words)}
\label{alg:HLF3}
\end{algorithm}
\begin{algorithm}
\SetAlgorithmName{Labeling Function}{}{}
\SetAlgoLined
\hspace*{\algorithmicindent} \textbf{Input:} Image, $n$: Number of classes, Kernels from first layer of pre-trained AlexNet (trained on ImageNet)\\
\hspace*{\algorithmicindent} \textbf{Output:} Probabilistic label vector\\
\tcc{\footnotesize{Deep learning based labeling function}}
m = Number of kernels from first layer of pre-trained AlexNet\;
\For{i=1$\cdots$ n}{
	\For{j = 1 $\cdots$ m}{
        $v_{avg_{ij}}$ = average value of Frobenius norm of activation map of $j^{th}$ kernel on subset of training samples from class $i$\;
    }
}
\For{j = 1 $\cdots$ m}{
        $v_{Image_{j}}$ = value of Frobenius norm of activation map of $j^{th}$ kernel on Image\;
    }

return OneHot($\argminA_i [\textbf{v}_{avg_{i}}\cdot \textbf{v}_{Image}]$)
\caption{Sample deep learned feature-based labeling function}
\label{alg:HLF4}
\end{algorithm}
\subsection{Ablation Studies with Labeling Functions}
\label{subsec_lf_ablation_study}
In order to understand the effect of different kinds of labeling functions, we performed an ablation study on the CIFAR10 dataset (considering it is the most natural of the considered datasets, and that it allows us to compute an Inception score to quantitatively compare the performance of various methods). In this study, we did not alter any hyperparameters described in Section \ref{sec_expts}. Our ablation study considers the following models:

\begin{enumerate}[label=M\arabic*:]
\item \textit{\model:} Full model
\item \textit{\model with no dependencies:} Same model as \model having 55 labeling functions, as in Table \ref{table:labeling_function_stat}. Each labeling function is, however, considered \textit{independent} of the other. 
\item \textit{\model with only heuristic labeling functions:} Same model as \model with 36 heuristic labeling functions but without any image processing or deep learned feature-based labeling functions
\item \textit{\model with only image processing labeling functions:} Same model as \model with only 17 image processing-based labeling functions but without heuristic or deep learned feature-based labeling functions
\item \textit{\model with only deep learned feature-based labeling functions:} Same model as \model with only 2 deep learned feature-based labeling functions but without heuristic or image processing labeling functions
\item \textit{\model with (heuristic labeling functions + deep learned feature-based labeling functions)}
\item \textit{\model with (deep learned feature-based labeling functions + image processing labeling functions)}
\item \textit{\model with (image processing labeling functions + heuristic labeling functions)}
\end{enumerate}
\begin{table}[]
\centering
\scalebox{0.85}{
\begin{tabular}{|c|c|c|c|c|c|c|c|c|}
\hline
 & M1 & M2 & M3 & M4 & M5 & M6 & M7 & M8 \\ \hline
\begin{tabular}[c]{@{}l@{}}Inception \\ Score\end{tabular} & \textbf{8.7} & 4.32 & 5.52 & 4.91 & 4.73 & 7.01 & 7.52 & 7.27 \\ \hline
\end{tabular}
}
\caption{Ablation study w.r.t labeling functions, as described in Section \ref{subsec_lf_ablation_study}}
\label{table:ablation_study}
\end{table}
The inception scores for the aforementioned 8 models are presented in Table \ref{table:ablation_study}. The base \model model comprising of all labeling functions outperforms all other models, highlighting the usefulness of a variety of labeling functions to model the non-trivial $P(\textbf{x}, y)$ distribution.
\section{More Qualitative Results}
\begin{figure*}
\centering
\includegraphics[width=1.05\textwidth]{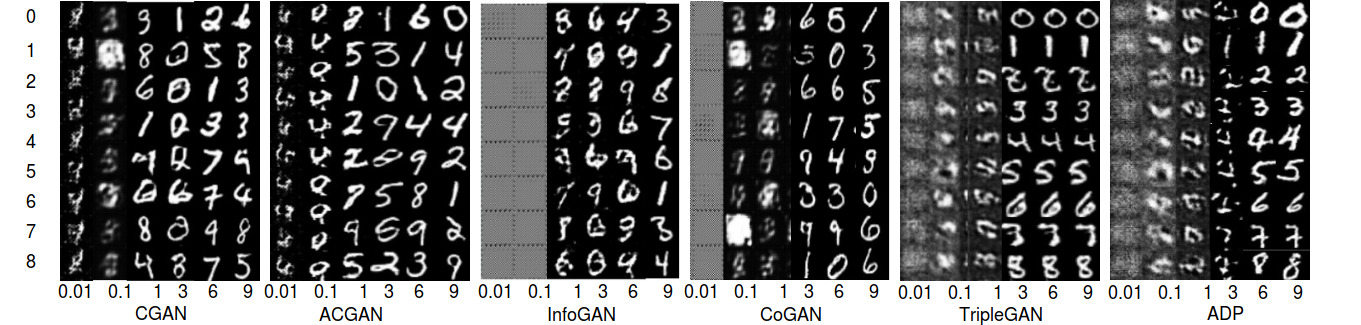}
\caption{Image-label pairs generated by training on MNIST dataset using CGAN, ACGAN, InfoGAN, CoGAN, TripleGAN and our method, ADP . For a given model, the columns of images represents generations after 0.01k, 0.1k, 1k, 3k, 6k and 9k epochs, and the rows correspond to the associated class label. It is evident that from 6k epochs onward, \model model starts generating quality images across classes and with a good image-to-label correspondence.}
\label{fig:MNIST_all}
\end{figure*}

\begin{figure*}
\centering
\includegraphics[width=\textwidth]{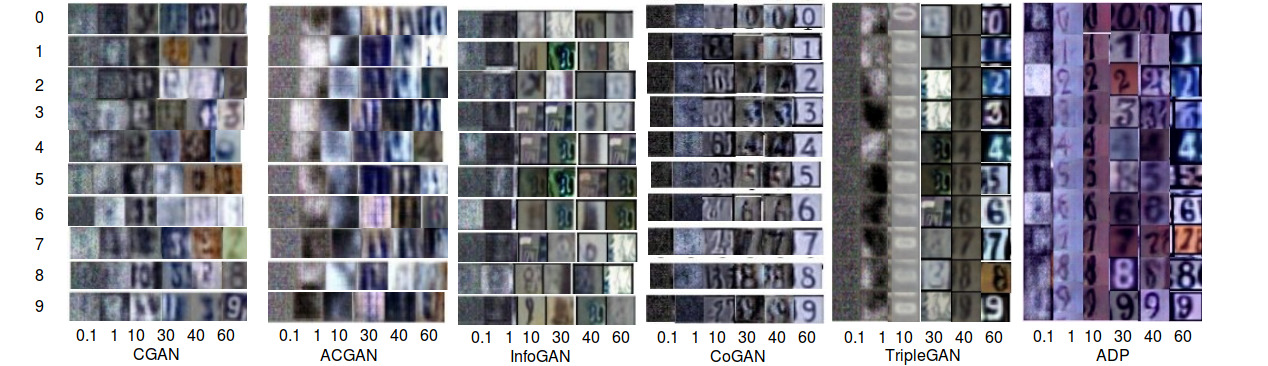}
\caption{Image-label pairs generated by training on SVHN dataset using CGAN, ACGAN, InfoGAN, CoGAN, TripleGAN and \model. For a given model, the columns of images represents generations after 0.1k, 1k, 10k, 30k, 40k and  60k epochs, and the rows correspond to the associated class label.}
\label{fig:SVHN_all}
\end{figure*}

\begin{figure*}
\centering
\includegraphics[width=1.05\textwidth]{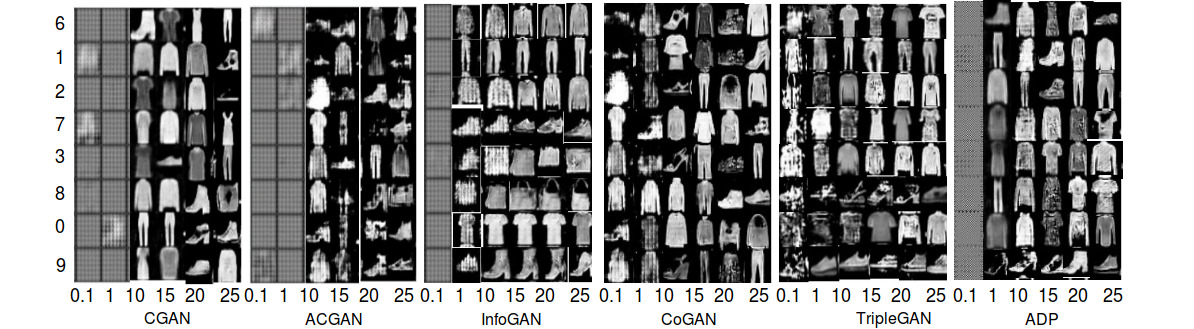}
\caption{Image-label pairs generated by training on Fashion MNIST dataset using CGAN, ACGAN, InfoGAN, CoGAN, TripleGAN and our method, ADP . For a given model, the columns of images represents generations after 0.1k, 1k, 10k, 15k, 20k and 25k epochs, and the rows correspond to the associated class label.}
\label{fig:FMNIST_all}
\end{figure*}

In addition to the results on CIFAR 10 presented in Section \ref{subsec_compare}, we also studied the performance of our \model method against other generative methods (CGAN, ACGAN, InfoGAN, CoGAN, TripleGAN) on MNIST, SVHN and Fashion MNIST datasets. Similar to CIFAR 10 generation, we changed the use case setup of the other methods to generate labeled images, using the publicly available code for each of the methods. Figures \ref{fig:MNIST_all}, \ref{fig:SVHN_all} and \ref{fig:FMNIST_all} present these results.

\paragraph{MNIST:} Figure \ref{fig:mnist_all} shows that both our method \model and TripleGAN generate good quality images on the MNIST dataset. We observe that both \model and TripleGAN give a high image-to-label correspondence. Surprisingly, state-of-the-art methods such as CGAN, ACGAN, InfoGAN and CoGAN fail to capture image-to-label correspondence despite generating good quality images. 
\paragraph{SVHN:} As shown in Figure \ref{fig:svhn_all}, we observe that our method generates human-recognizable images with a good image-to-label correspondence in just 1k epochs on the relatively harder SVHN dataset. At higher epochs, CoGAN (epoch = 30) and TripleGAN (epoch = 40) also generate images of good quality, but broadly fail to capture different styles, backgrounds and illuminations of the generated digit.
\begin{figure*}
\centering
\includegraphics[width=\textwidth]{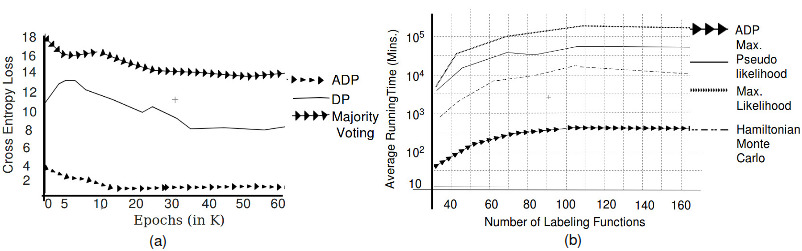}
\caption{(a) Test-time classification cross-entropy loss of a pre-trained ResNet model on image-label pairs generated by ADP, \model (i.e. only its Image-GAN component) with majority voting and \model (i.e. only its Image-GAN component) with DP for labels; (b) Average running time of \model against other methods to estimate the relative accuracies and inter-function dependencies in DP.}
\label{fig:Cross}
\end{figure*}
\begin{table*}
\centering
\begin{tabular}{|l|l|l|l|l|l|l|l|}
\hline \hline
 & GAN & CGAN & ACGAN & InfoGAN & CoGAN & ADP & Triple\\ \hline \hline
MNIST & 198 & 201 & 204 & 225 & 278 & \textbf{344} & 321 \\ \hline
FMNIST & 213 & 206 & 234 & 276 & 254 & 292 & \textbf{312}\\ \hline
SVHN & 87 & 145 & 178 & 158 & 123 & \textbf{246} & 223\\ \hline \hline
\end{tabular}
\caption{Parzen window based evaluation on MNIST, FMNIST and SVHN datasets.}
\label{table:parzen_window}
\end{table*}
\paragraph{Fashion MNIST:} Most of the considered methods do well on this dataset. \model and TripleGAN provide the sharpest results on close visual observation.  
\section{More Quantitative Results}
\paragraph{Parzen Window Based Evaluation:}
In addition to the results with Inception score and \textit{HTT} presented in Section \ref{subsec_compare}, we compared our method against other generative models (described in Section \ref{sec_expts}) based on the Parzen window score at test time. Parzen window \cite{breuleux2010unlearning} is a commonly used non-parametric density estimation method to evaluate generative models (especially GANs \cite{goodfellow2014generative}) for which exact likelihood is not tractable. Based on the samples generated by the model, we use a Parzen window with a Gaussian kernel as a density estimator. This helps obtain a proxy for true log-likelihood and thereby evaluate test log-likelihood. These results are shown in Table \ref{table:parzen_window}. The table shows that \model has performed significantly well on MNIST (Score is 344) and SVHN (Score is 246) dataset and outperformed other state-of-the-art models including TripleGAN. For Fashion MNIST, our method is a close second with respect to TripleGAN. We chose the Parzen window size using cross-validation, as described in \cite{goodfellow2014generative}. 
\begin{figure*}[h]
\centering
\includegraphics[width=0.7\textwidth]{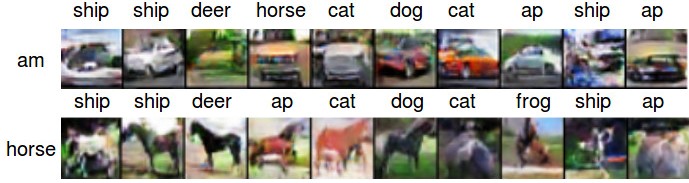}
\caption{Sample results of image-label pairs generated by combining a vanilla GAN (for image generation) and DP \cite{ratner2016data} (for label generation) using the same labeling functions used in this work. Row labels represent the original class label (am = automobile) and column labels are provided by DP. Note the poor image-label correspondence, supporting the need for our work.}
\label{fig:DP_Label}
\end{figure*}
\section{More Results on Multi-task Joint Distribution Learning}
\begin{figure}
\centering
\includegraphics[width=0.5\textwidth]{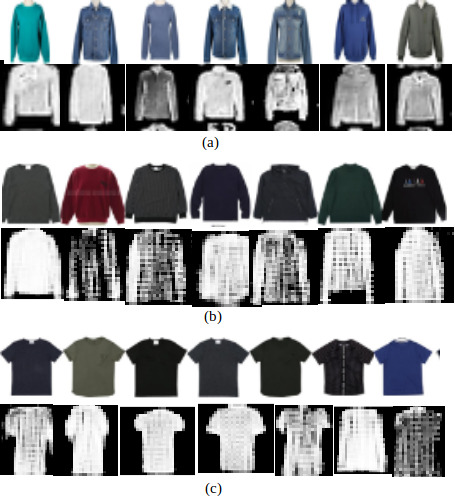}
\caption{Demonstration of cross-domain multi-task learning using \model: (a)(b) Generated samples of Shirt (class 6 of Fashion MNIST dataset); (c) Generated samples of T-shirt (class 0 of Fashion MNIST dataset). Samples generated of the LookBook dataset are color images (top rows), while those of Fashion MNIST are grayscale images (bottom rows).}
\label{fig:Transfer_all}
\end{figure}

In continuation to the results presented in Section \ref{sec_analysis} (and \ref{fig:Motivational_example}), we present more results for the capability of \model to perform multi-task joint distribution learning in 
Figure \ref{fig:Transfer_all}. The figure captures our promise and shows that \model is able to generate samples from two different domains, including samples of different colors. 

\section{Comparison against Vote Aggregation Methods}
\paragraph{Comparison against Majority Voting and DP:}
To study the usefulness of learning relative accuracies and inter-function dependencies using \model, we compared the performance of our method, both with majority voting and Data Programming (DP, \cite{ratner2016data}). In majority voting, $LFB$ does not estimate relative accuracies and inter-function dependencies of labeling functions as described in Section \ref{sec_adp_methodology}. Instead, for a given image, each labeling function of $LFB$ makes a probabilistic prediction, and we take a maximum vote to obtain the final label. As in Section \ref{subsec_compare}, we studied the test-time classification cross-entropy loss of a pre-trained ResNet model on image-label pairs generated by ADP, \model (i.e. only its Image-GAN component) with majority voting and DP. The results are presented in Figure \ref{fig:Cross}a, which shows that \model has significantly lower cross-entropy loss than the other two methods, thus corroborating its effectiveness.

\paragraph{Adversarial Data Programming vs MLE-based Data Programming:}
To further quantify the benefits of our \model, we also show how our method compares against Data Programming (DP) \cite{ratner2016data} using different variants of MLE: MLE, Maximum Pseudo-likelihood, and Hamiltonian Monte Carlo. We note that DP only aggregates labels; we hence, combined a vanilla GAN with DP as separate components to conduct this study. We started with a small number of labeling functions (viz., 35 functions) and progressively added additional labeling functions, noting the time taken by each aforementioned parameter estimation method. Figure \ref{fig:Cross}b presents the results and shows that \model is almost 100X faster than MLE-based estimation. Figure \ref{fig:DP_Label} also shows sample images generated by the vanilla GAN, along with the corresponding label assigned by MLE-based DP using the same labeling functions as used in our work. Clearly, the labels are incorrect, thus supporting the value of the proposed work in learning a joint distribution, than combining two individual components.


\end{document}